\documentclass{article}
\usepackage{ijcai19}

\usepackage{times}
\usepackage{soul}
\usepackage{url}
\usepackage[hidelinks]{hyperref}
\usepackage[utf8]{inputenc}
\usepackage[small]{caption}
\usepackage{graphicx}
\usepackage{amsmath}
\usepackage{booktabs}
\usepackage{subfigure}
\usepackage{bm}
\usepackage{algorithm}
\usepackage{algorithmic}
\title{Reinforcement Learning Experience Reuse with Policy Residual Representation\footnote{This work is supported by the National Key R\&D Program of China (2018YFB1004300), NSFC (61751306, 61876077), Jiangsu SF (BK20170013), and Collaborative Innovation Center of Novel Software Technology and Industrialization. This work was partially done while Wen-Ji Zhou was interning at Fuxi AI Lab. Yang Yu is the corresponding author.}}

\author{
Wen-Ji Zhou$^1$\and
Yang Yu$^1$\and
Yingfeng Chen$^2$\and
Kai Guan$^2$\and\\
Tangjie Lv$^2$\and
Changjie Fan$^2$\and
Zhi-Hua Zhou$^1$\\
\affiliations
$^1$National Key Laboratory for Novel Software Technology, Nanjing University, Nanjing, China\\
\emails \{zhouwj, yuy, zhouzh\}@lamda.nju.edu.cn,\\
$^2$NetEase Fuxi AI Lab, Hangzhou, China\\
\emails
\{chenyingfeng1,guankai1,hzlvtangjie,fanchangjie\}@corp.netease.com
}

\begin{document}

\maketitle

\begin{abstract}
Experience reuse is key to sample-efficient reinforcement learning. One of the critical issues is how the experience is represented and stored. Previously, the experience can be stored in the forms of features, individual models, and the average model, each lying at a different granularity. However, new tasks may require experience across multiple granularities. In this paper, we propose the policy residual representation (PRR) network, which can extract and store multiple levels of experience. PRR network is trained on a set of tasks with a multi-level architecture, where a module in each level corresponds to a subset of the tasks. Therefore, the PRR network represents the experience in a spectrum-like way. When training on a new task, PRR can provide different levels of experience for accelerating the learning. We experiment with the PRR network on a set of grid world navigation tasks, locomotion tasks, and fighting tasks in a video game. The results show that the PRR network leads to better reuse of experience and thus outperforms some state-of-the-art approaches.
\end{abstract}

\section{Introduction}
Reinforcement learning \cite{SuttonB98} has recently shown many impressive results.
It has achieved human-level performance in a wide variety of tasks, including playing Atari games from raw pixels \cite{GuoSLLW14,Mnih15,SchulmanLAJM15}, playing the game of go \cite{SilverHMGSDSAPL16} and robotic manipulation \cite{LevineFDA16,LillicrapHPHETS15}.
However, most of them have very high sample complexity.
For example, to master a single Atari game, state-of-the-art methods need tens of thousands of experience to learn a policy.
However, while facing a new task, the same quantity of samples is needed in traditional reinforcement learning methods.
In contrast, a human can learn good policy through limited trials on a new task, when they have a similar experience.
It's essential to utilize knowledge learned from experienced tasks, in order to achieve efficient reinforcement learning.

Various ways have been investigated to enable reinforcement learning with the ability of experience reuse. In the process of experience reuse, the first step is how the experience is represented and stored. The experience can be stored in the forms of features, individual models, and an average model. In, e.g., \cite{BarretoDMHSSH17,AmmarTTDW12}, the experience is stored as the features learned from the old tasks. In, e.g., \cite{YuCDZ18,HuYZ18,ZhangIJCAI18}, the experience is stored in individual models, each of which was trained on an old task. The reuse is to find a combination of the individual models on the new task. In, e.g., \cite{FinnAL17,GuptaMLAL18}, the experience is stored in an average model, where the average was measured as the gradient update steps to reach the optimal model over the old tasks. The reuse is to start with the model and continue the training on the new task. We notice that each previous method represents and stores the experience in a particular granularity, which does not only restrict the effectiveness of the reuse, but also make assumptions on the relationship between the new tasks and the old tasks. It could be useful for a very similar task to reuse samples directly, but for a dissimilar task, only the average model could be helpful. Therefore, for a more practical method of experience reuse, it could be beneficial that the experience is represented in multiple levels. 


In this work, we propose a multi-level model architecture, named as \emph{policy residual representation} (PRR), as well as the training method that enables a single model to represent multiple levels of experience. In each level of PRR, there are one or more component modules, corresponding to a subset of the tasks. The training starts from the top level with one module corresponding to all the tasks, i.e., the module is the policy that maximizes the sum of (normalized) reward on all the tasks. In other words, the top level learns the average policy over all the tasks. In each following level, a module is learned over a selected subset of tasks according to a predefined mask. Moreover, when learning the module, all the upper levels are fixed, and thus the module learns a residual policy over the selected tasks. In this way, PRR forms a multi-level architecture, where the experience of different granularities can be represented. When learning in a new task, the reuse of the PRR experience can have two phases: the first is to select the experience by learning the weights of the modules in all the levels, then a new module can be appended to learn the residual policy specific to the new task. 

We test PRR with the reinforcement learning algorithm PPO \cite{Schulman17PPO} on three sets of environments, a set of grid-world navigation tasks, a set of Mujoco \cite{TodorovET12} controlling tasks, and a set of fighting tasks in a video game. Experiment results not only show that PRR leads to superior performance with experience reuse to some state-of-the-art methods, but also verify the applicability of PRR to various tasks.

\section{Background}

\noindent{\bf Reinforcement learning} aims at learning the optimal policy model from autonomous interactions with the environment. 

The reinforcement learning problem can be expressed as a Markov decision process. In this paper, we adopt the standard deep reinforcement learning setting. $S$ denotes the state set of the environment and $A$ denotes the action set of an agent. The agent interacts with an environment over periods of time according to a behavior policy $\pi_{\theta}$.
We usually use a neural network to learn a policy $\pi_{\theta}$ and $\theta$ denotes its parameters.
At each time step $t$, the agent gets a state observation $s_t \in S$ from environments.
Then it samples an action $a_t\sim\pi_{\theta}(s_t)$ and applies the action to the environment.
The environment then returns a reward $r_t$ sampled from an unknown reward function $R(s_t,a_t)$ to the agent and shifts to a new state $s_{t+1}$ from an unknown transition function $P(s_{t+1}|s_t,a_t)$.
Meanwhile, the environment will notify the agent whether to terminate the current episode.
The goal of agent is to maximize the expected future discounted reward $\eta(\pi_{\theta})=E_\tau[\sum_{t=0}^T \gamma^t r_t]$, where $\tau$ denotes the whole trajectory.

However, sampling inefficiency is one of the main difficulty in reinforcement learning.
For example, the state-of-the-art methods need tens of thousands of experience to learn a policy on a video game.
Especially when the dimension of the state is too high or the environment is complex, the sampling efficiency of traditional reinforcement learning is very limited.\\[0pt]

\noindent{\bf Experience reuse} from old tasks is an effective way to improve the sampling efficiency of reinforcement learning on new tasks. The first step is to represent and store the experience.
Some approaches \cite{BarretoDMHSSH17,AmmarTTDW12} store the experience as the features learned from the old tasks.
Some approaches \cite{YuCDZ18,HuYZ18,ZhangIJCAI18} store the experience in individual models. Each of the individual models is trained on an old task. The experience can be reused by finding a combination of the individual models on the new task.
And another kind of approaches, e.g. MAML \cite{FinnAL17} and MAESN \cite{GuptaMLAL18}, store the experience in an average model. The average was measured as the gradient update steps to reach the optimal model from the old tasks.
The reuse is to start with this average model and continue training on the new task.

However, each previous method represents and stores the experience in a particular granularity.
The fixed granularity will restrict the effectiveness of the reuse and make assumptions on the relationship between the new tasks and the old tasks.
These methods have been shown useful, for learning new tasks with certain types of similarity with the old tasks. However, they may fail for tasks with a different similarity type that requires a different level of experience.

Different from the existing methods, this paper presents the policy residual representation (PRR) network, which aims at extracting multiple levels of experience by learning policy residual representations with its multi-level structure.

\section{Policy Residual Representation Network}

\subsection{PRR architecture}
Policy residual representation (PRR) is a multi-level neural network architecture. But unlike multi-level architectures in hierarchical reinforcement learning that are mainly used to decompose the task into subtasks, PRR employs a multi-level architecture to represent the experience in multiple granularities.


In our setting, we have a set of experience tasks $E=\{e_1, e_2, \ldots, e_n\}$, which share the same state space and action space. A PRR model is to be trained in these environments, and storqes the experience in different granularities. The overall idea of PRR is illustrated in Figure \ref{fig:structure}. Firstly, we show the structure of PRR network in Figure \ref{fig:structure} (a). PRR is a multi-layer network structure, each layer has one or more modules. $L_{ij}$ denotes the $j$-th module in the $i$-th level. Each module receives a state as input and outputs a component of the action. The outputs are all summed up according to the linear weights $\bm{w}$ to obtain the final action distribution, i.e., 
\begin{equation}
\bm{a} = w_0\bm{a}_0 + w_{11}\bm{a}_{11} + ... + w_{ij}\bm{a}_{ij}.
\label{eq:weighted sum}
\end{equation} 
We restrict that $\bm{w}$ is a unit vector, i.e., $||\bm{w}||_1=1$. Also, note that it is not straightforward to define what is a ``negative'' task, therefore, we require that each $w$ is none negative, i.e., $w_{ij} \in [0,1]$.
Each module learns a representation of policy residuals. Thus this structure is called policy residuals representation network.


\subsection{Experience acquisition with PRR model}
In order to learn the experience from different granularity, we train the modules in a PRR sequentially, in a top-down manner. First, we need to prepare the masks, which specify the tasks selected for the modules. The multi-granularity of the experience representation has a nature similarity of the spectrum representation, thus it is natural to think of the Walsh functions \cite{Walsh} to generate the masks, which achieves a spectrum transformation. However, Walsh functions generate codes with $1$'s and $-1$'s, but it is unnatural to define a negative reinforcement learning task. For example, while ``achieving the goal'' can correspond to an optimal policy, ``not achieving the goal'' corresponds to infinite polices. Therefore, a possible modification from the Walsh functions can be that we separate $1$'s and $-1$'s as different sets of tasks, each is used to train a module. Other ways to define the masks are possible, however, in this paper, we do not focus on finding the best masks. We focus on validating the idea of PRR. Therefore, we set a mask $m_{ij}$ as a binary vector, which has the same length with the number of tasks. Being 1 in the mask means the corresponding task is included for the module, and 0 means excluded. 

\begin{figure}[t!]
\centering
\includegraphics[width=\columnwidth]{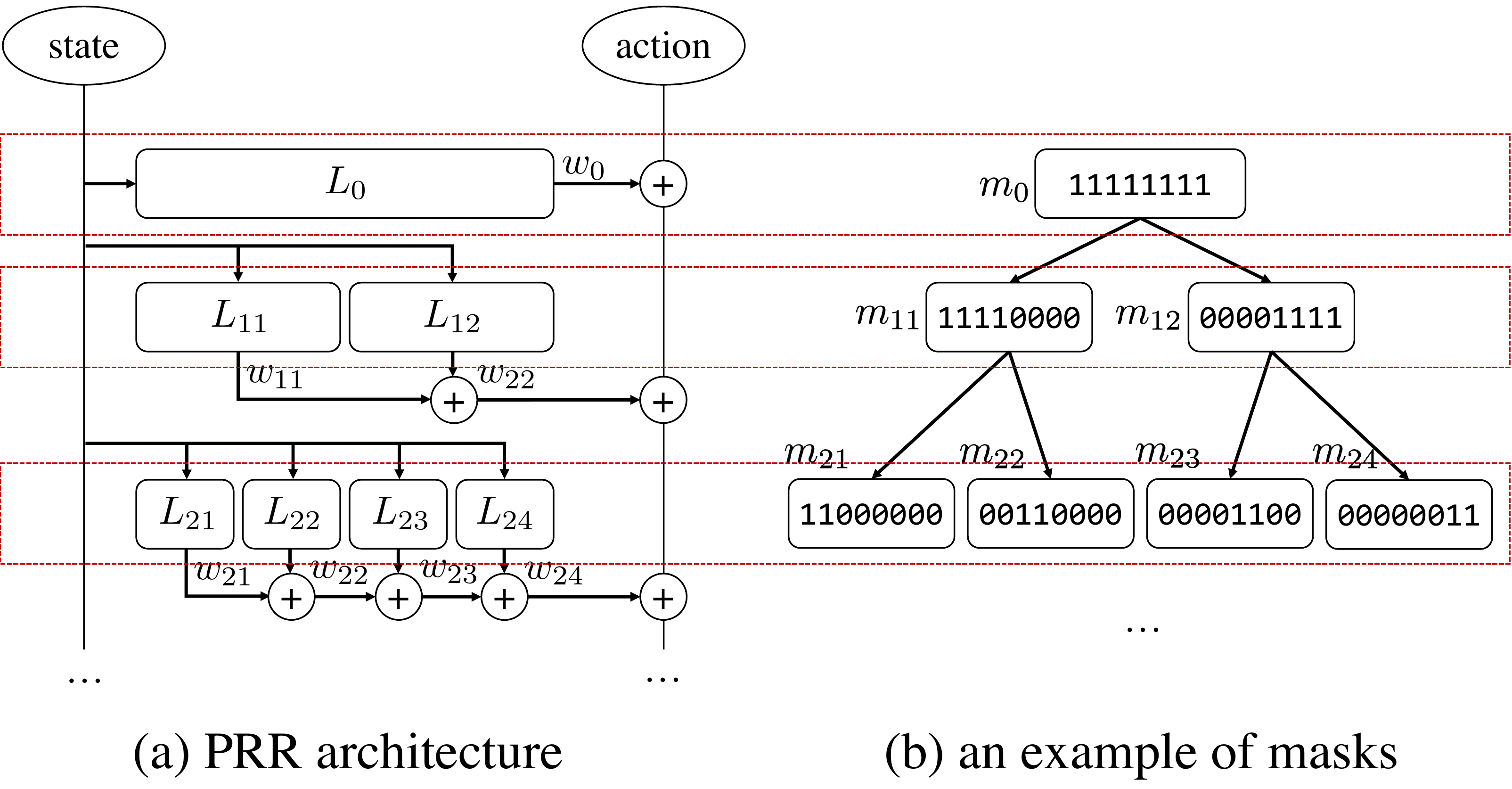}
\caption{Architecture of PRR. (a) the multi-level structure of PRR, where $L_{ij}$ denotes the $j$-th module in the $i$-th level, and the modules' output are summed up with weights $w_{ij}$ to achieve the final action (b) an example of hierarchical masks that specify the task subset for each module.}
\label{fig:structure}
\end{figure}

The masks used in this paper are shown in Figure \ref{fig:structure} (b).
The mask $m_{ij}$ is a binary vector, which has the same length with the number of tasks.
Each dimension corresponds to a task.
$m_{ij}^k=1$ indicates that we select $k$-th task into training task sets for $L_{ij}$.
When training $L_{ij}$, we uniformly sample a task from the selected task set and train $L_{ij}$ on this task. Then another task is uniformly sampled and $L_{ij}$ is trained. In this process, the parameters of other modules are frozen.


By learning on a subset of previous tasks, $L_{ij}$ extracts experience of a certain granularity.
For example, we usually let $L_0$ module learn the overall experience of this series of tasks, which is the most coarse-grained experience.
So its mask $m_0$ is a vector that all elements are 1.
If we set $m_{ij}$ to a zero-vector except for one dimension, the $L_{ij}$ will learn on a single task, by which $L_{ij}$ learns the most fine-grained experience.
So we use masks to train PRR from top to bottom, from coarse-grained to fine-grained, as shown in Figure \ref{fig:structure}. This experience acquiring process is shown in \ref{alg:train_PRR}.

\begin{algorithm}[t!]
\caption{module training of $L_{ij}$}
\label{alg:train_Lij}
\hspace*{0.02in} {\bf Input:}
Environments: $\{e_1,e_2,...,e_n\}$\\
\hspace*{0.42in} Previous network parameters: $\{\theta_0,\theta_{11},...,\theta_{ij-1}\}$\\
\hspace*{0.42in} Previous weights: $\{w_0,w_{11},...,w_{ij-1}\}$\\
\hspace*{0.42in} Mask: $m_{ij}$\\
\begin{algorithmic}[1]
\STATE Initialize $\theta_{ij}$ and let $w_{ij}=0$
\STATE Apply $m_{ij}$ to select a sub set $Envs$ from $\{e_1,e_2,...,e_n\}$
\STATE $\bm{w} = [w_0,w_{11},...,w_{ij}]$
\REPEAT
\STATE Random sample $e_i$ from $Envs$
\STATE $\hat{\bm{w}} = $normalize$(\bm{w})$
\STATE $\tau=\emptyset$
\WHILE{episode not terminate}
\STATE Get state $\bm{s}_t$ from $e_i$
\STATE Predict $A = [\bm{a}_0,\bm{a}_{11},...,\bm{a}_{ij-1}]$ from $\bm{s}_t$ using $[\theta_0,\theta_{11},...,\theta_{ij}]$
\STATE $\bm{a}_t = \hat{\bm{w}}^TA$
\STATE Act $\bm{a_t}$ and receive reward $r_t$
\STATE Add $(\bm{s}_t,\bm{a}_t,r_t)$ into $\tau$
\ENDWHILE
\STATE Apply PPO to update $\theta_{ij}$ and $\bm{w}$ using $\tau$
\UNTIL{convergence}
\RETURN $\{\theta_0,\theta_{11},...,\theta_{ij}\}$, $\{w_0,w_{11},...,w_ij\}$
\end{algorithmic}
\end{algorithm}

\begin{algorithm}[t!]
\caption{Experience acquiring with PRR model}
\label{alg:train_PRR}
\hspace*{0.02in} {\bf Input:}
Environments: $E=\{e_1,e_2,...,e_n\}$\\
\hspace*{0.42in} Masks: $M=\{m_0,m_{11},m_{12},...,m_{ij}\}$\\
\begin{algorithmic}[1]
\STATE Initialize network parameter set $P=\emptyset$
\STATE Initialize weight set $W=\emptyset$
\FOR{$m$ in $M$}
\STATE $P$,$W$ = Algorithm-\ref{alg:train_Lij}($E,P,W,m$)
\ENDFOR
\RETURN $P$, $W$
\end{algorithmic}
\end{algorithm}

\begin{algorithm}[t!]
\caption{Experience reusing with PRR model}
\label{alg:transfer_PRR}
\hspace*{0.02in} {\bf Input:}
Previous environments: $E=\{e_1,e_2,...,e_n\}$\\
\hspace*{0.42in} New Environments: $e_{new}$\\
\hspace*{0.42in} Existing network parameters: $P=\{\theta_0,\theta_{11},...,\theta_{ij}\}$\\
\hspace*{0.42in} Existing weights: $W=\{w_0,w_{11},...,w_{ij}\}$\\
\begin{algorithmic}[1]
\STATE Add $e_{new}$ to $E$
\STATE Let $m_{new}=\{0,0,...,1\}$
\STATE $P$,$W$ = Algorithm-2($E,P,W,m_{new}$)
\RETURN $P$, $W$
\end{algorithmic}
\end{algorithm}

To avoid the PRR network overfits a certain environment, we need to ensure that every task has the same importance.
So we scale rewards on different tasks to the same range.
We can apply a policy gradient based reinforcement learning approach to optimize parameters of $L_{ij}$ and $\bm{w}$.
We choose to apply PPO \cite{Schulman17PPO} in this paper.

\subsection{Experience reuse with PRR model}
When we want to reuse PRR to a new task, we can first try to recombine existing modules.
It means that we freeze parameters of all the modules and only train the weights of linear combination $\bm{w}$.
Because there are fewer parameters to be optimized, PRR can transfer to the new task rapidly.

However, if the new task is not so much similar to previous tasks, the experience stored in PRR maybe not enough to solve it well.
Thus a new module $L_{new}$ should be added to PRR and it learns policy residual representation on a new task.
Because PRR has learned the coarse-grained experience of this kind of tasks, we let the mask $m_{new}=\{0,0,...,1\}$.
It means that the selected task set for $L_{new}$ only contains the new task, which let $L_{new}$ learn the fine-grained experience on the new task.
During training, our approach only updates the parameters of $L_{new}$ and $w_{new}$ and freezes other parameters.
The learning process of transferring to a new task is shown in Algorithm \ref{alg:transfer_PRR}.

\begin{figure*}[ht]
  \centering
  \subfigure[Learning $L_0$]{
  \includegraphics[width=0.23\linewidth]{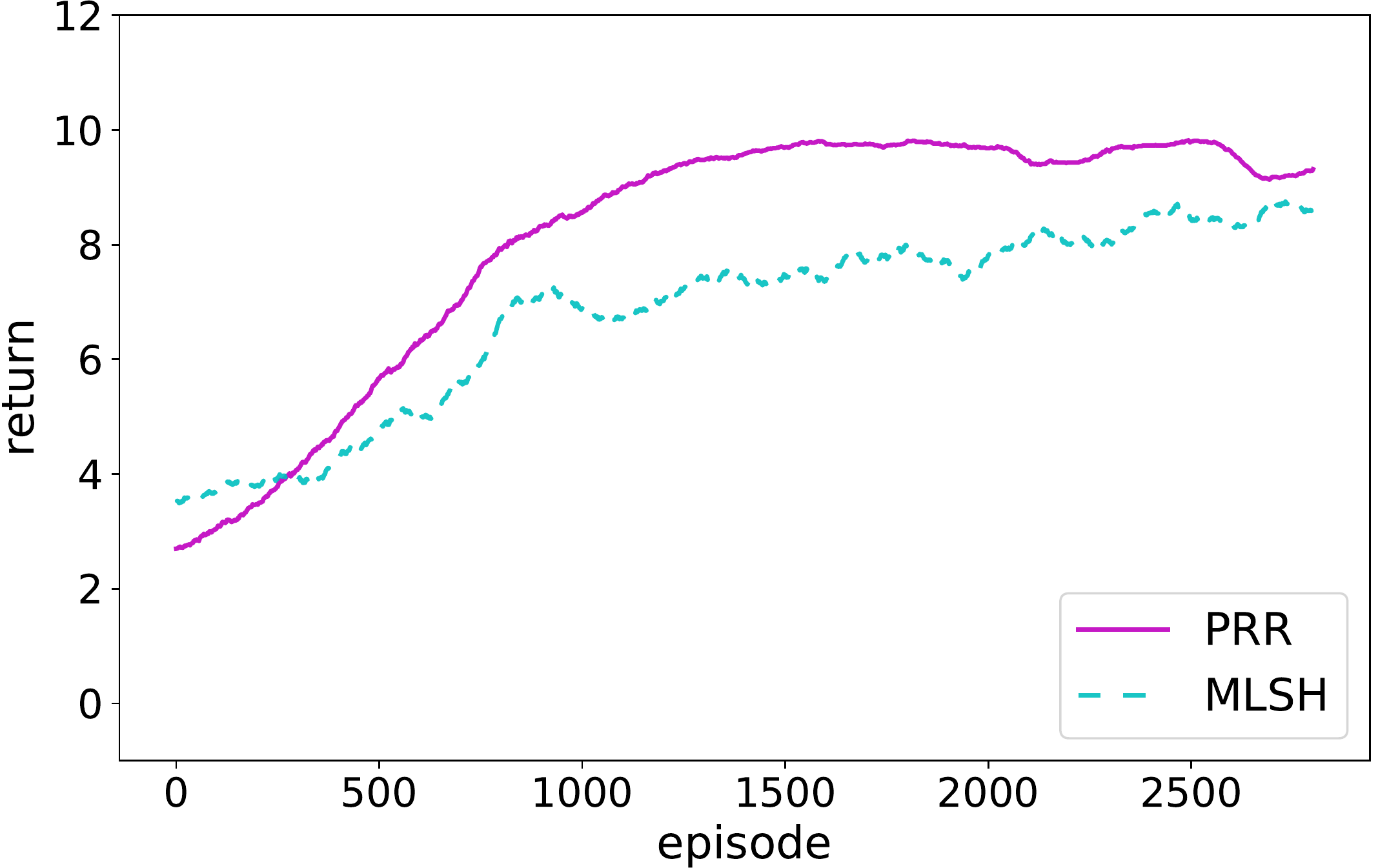}
  }
  \subfigure[Learning $L_{11}$]{
  \includegraphics[width=0.23\linewidth]{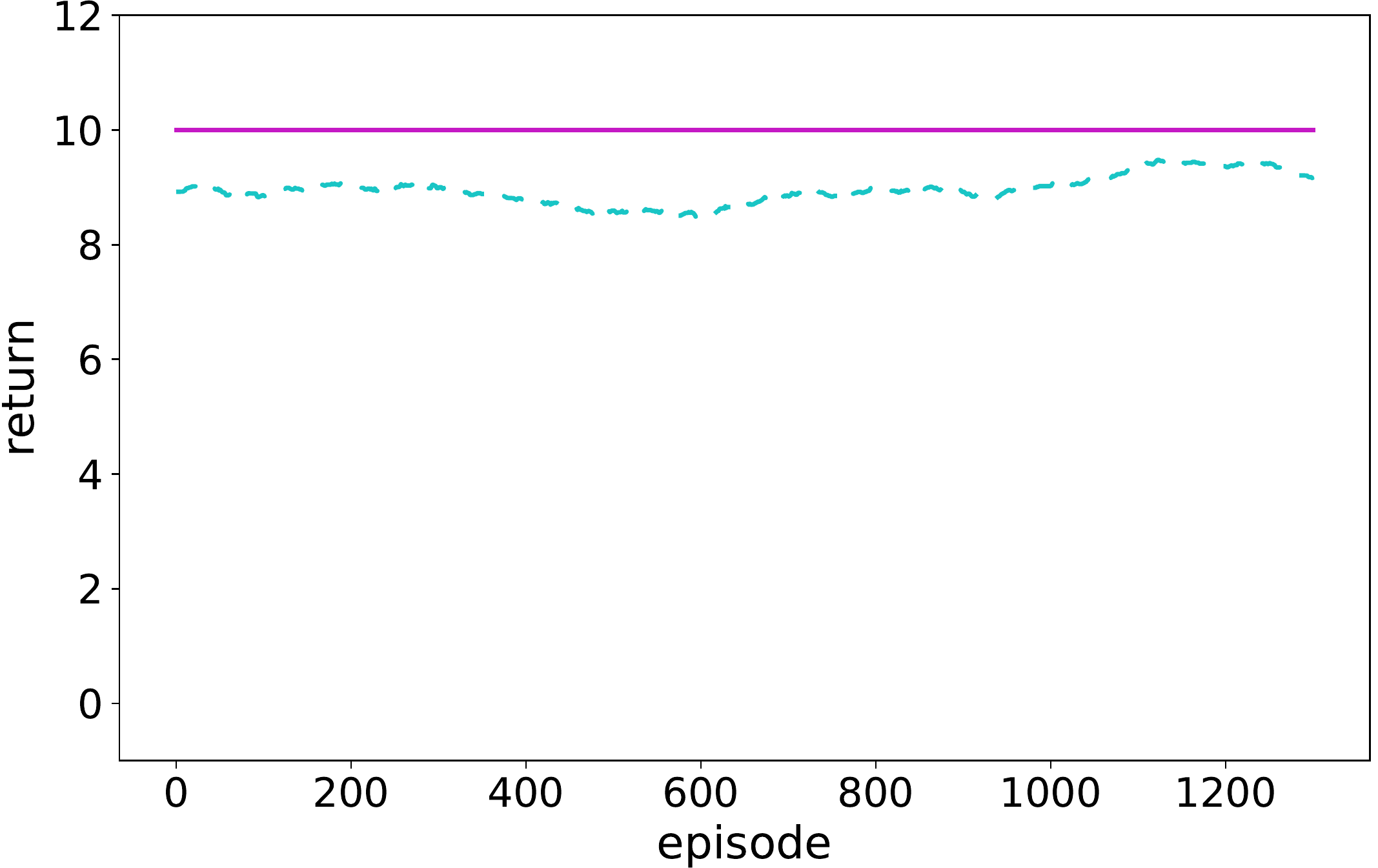}
  }
  \subfigure[Learning $L_{12}$]{
  \includegraphics[width=0.23\linewidth]{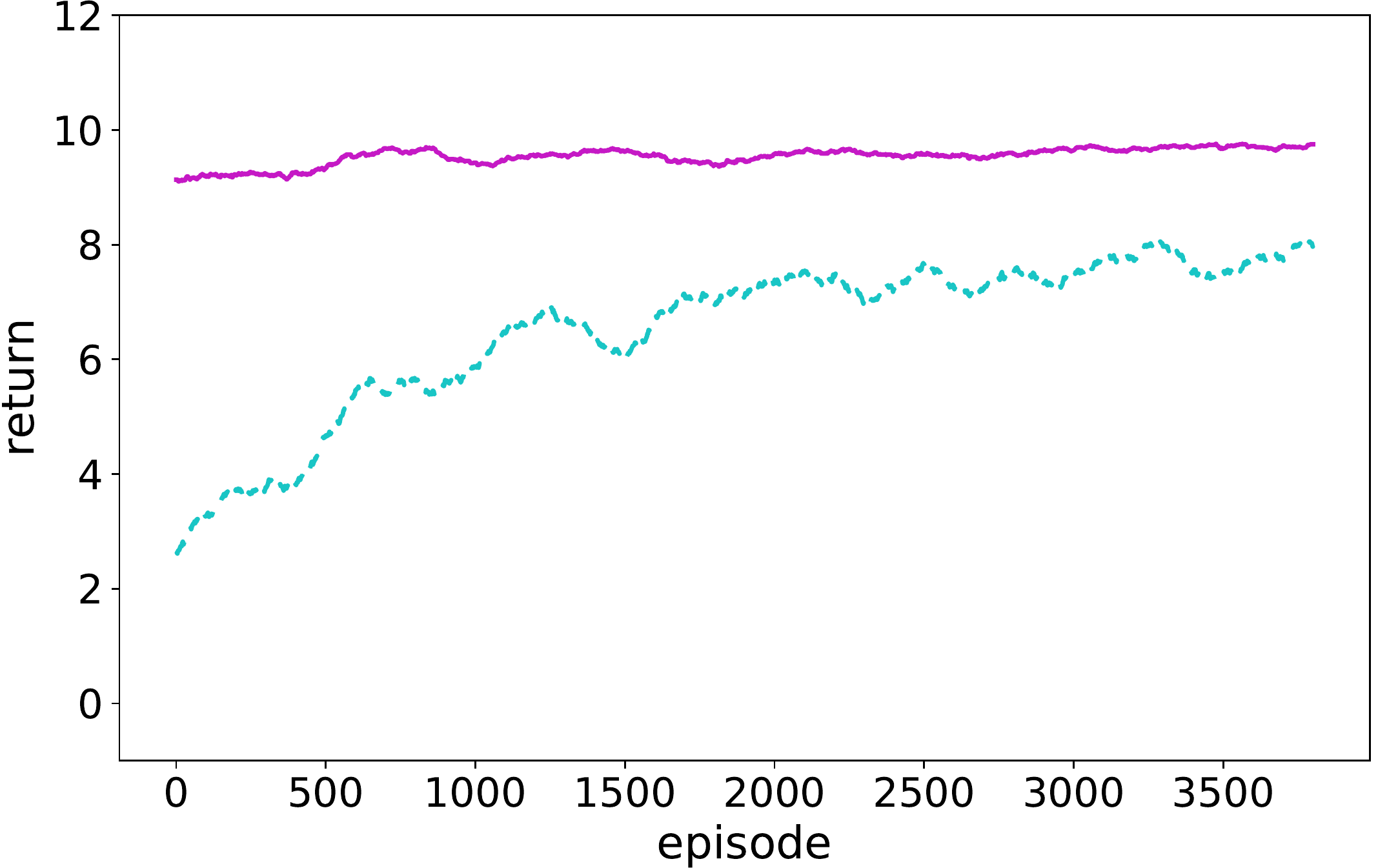}
  }
  \subfigure[Learning $L_{13}$]{
  \includegraphics[width=0.23\linewidth]{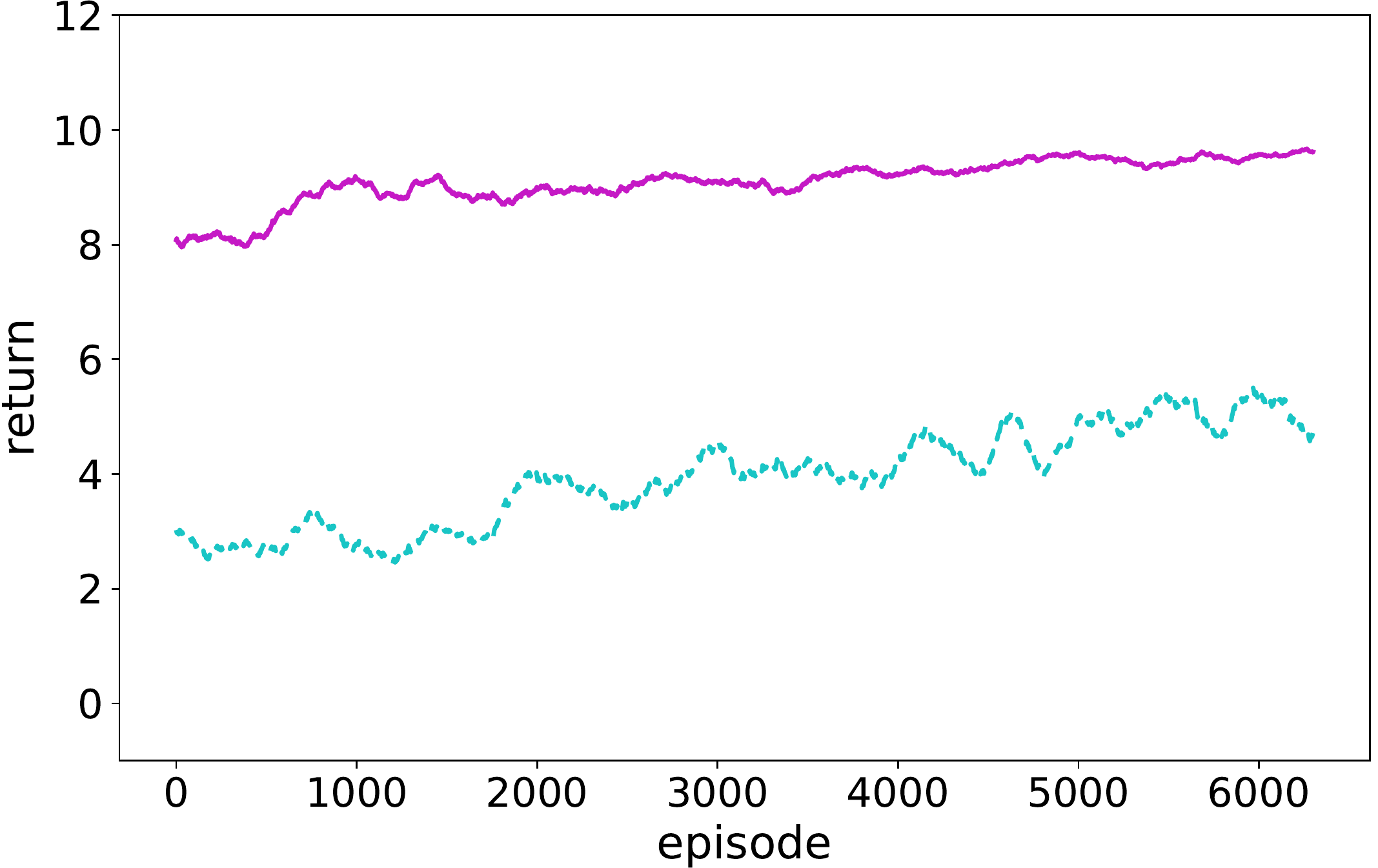}
  }
  \caption{The training process of PRR and the comparison to MLSH.}
  \label{fig:keyEnvLearnProcess}
  \end{figure*}
  
\section{Experiments}
We conduct several experiments to demonstrate that PRR can learn different levels of experience from old tasks and reuse the experience to learn on a new task, by learning different levels of policy residual representation.
In all these experiments, we use PPO as a basic reinforcement learning algorithm of PPR.
We train PRR on level 0 and level 1 in all these experiments.
On level 0, as mentioned before, we train $L_0$ to learn the overall experience of this series of tasks, which is the most coarse-grained experience.
On level 1, we learn policy residual networks from $L_{11}$ to $L_{1n}$, where $n$ is the number of environments.
The mask of $L_{1j}$ is a zero vector except the $j$-th element, which means that we train $L_{1j}$ only on $j$-th environment.

We compare our method with basic PPO \cite{Schulman17PPO}, MLSH \cite{frans2017meta}, MAML\cite{FinnAL17} and MAESN\cite{GuptaMLAL18}.
Basic PPO is a baseline algorithm that is directly trained in a new environment.
MLSH, MAML, and MAESN are Meta-RL algorithms.
They all focus on multi-task reinforcement learning.
MLSH also uses a multi-layer structure.
It learns common low-level skills from a series of similar environments and reuses them on new environments.

\begin{figure}
\centering
\includegraphics[width=2.6in]{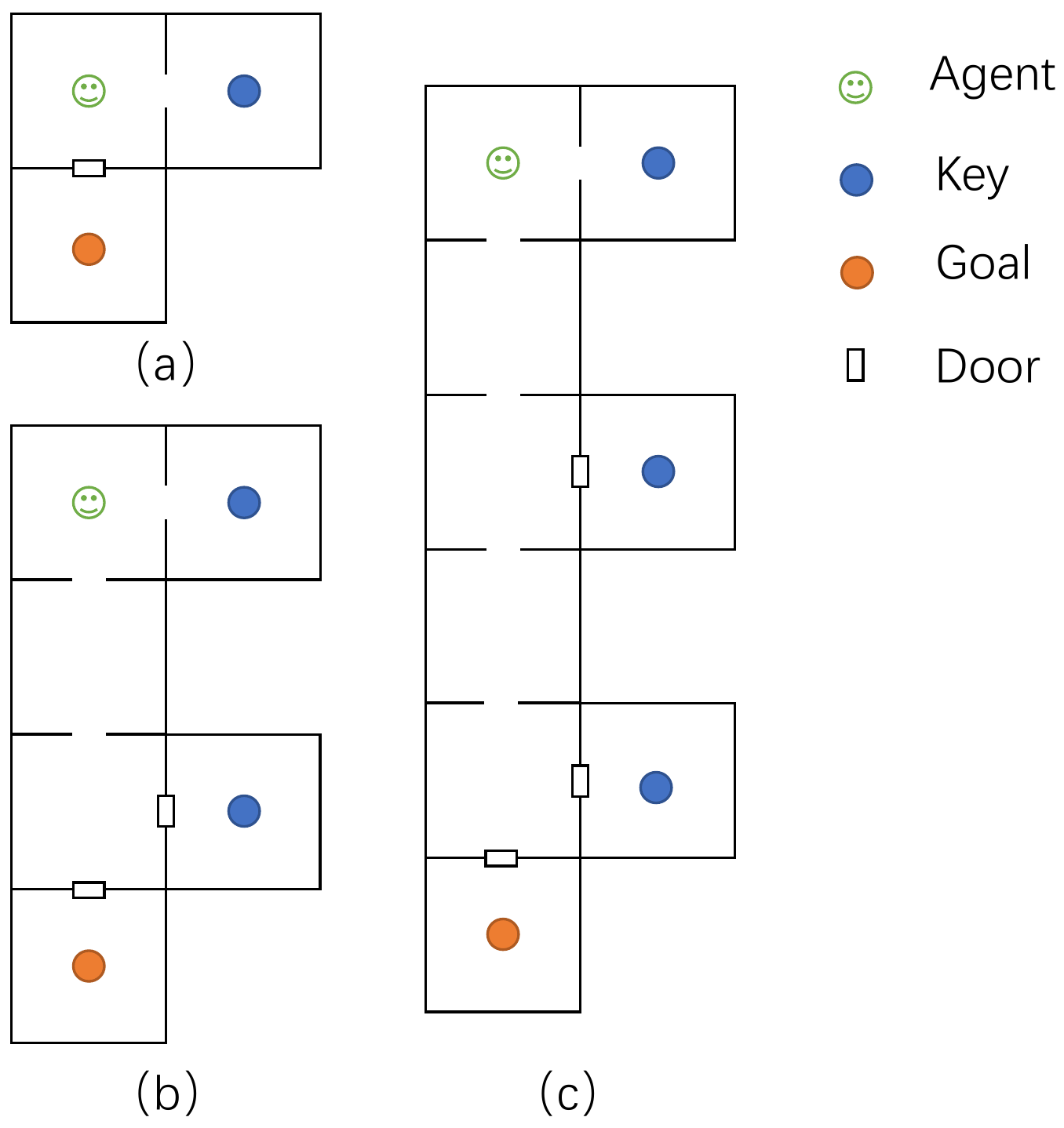}
\caption{FetchTheKey environments. The agent is born randomly in the up-left room, and the keys are randomly placed in their rooms, and the goal position is fixed.}
\label{fig:FTK_env}
\end{figure}

We have several experiments to test the performance of PRR network.
We also analyze the necessity of learning from different granularities of experience and the importance of learning new policy residual representations in the new environment.
All comparison algorithms use the same hyperparameters in the PPO algorithm.
The results of all experiments are the average of 10 repeated experiments.

\subsection{Complex Grid World Environment}
In order to test the performance and analyze the details of the learning process, we first validate our architecture on a complex grid world navigation tasks, which is called FetchTheKey tasks.

FetchTheKey environment is a grid world environment with many rooms, as shown in Figure \ref{fig:FTK_env}.
There are many rooms in this environment.
Some rooms are connected by doors and doors are initially locked.
The agent can open the door only if it has the right key.
The room that has the first key can be entered freely.
After that, the agent needs key $i$ to enter the room that has key $i+1$.
And if the agent has all the keys, it can enter the final room to reach the goal.
The positions of keys are randomly initialized every time we reset the environment, while the position of the goal is fixed.
The agent can observe its position, the number of keys it picked and the grid information around it with a radius of 2.
It means that the observation has 27 dimensions.
The agent can take 4 actions, which are four moving directions.
It will pick keys automatically when it is in the same position with a key.
The size of the rooms is $5\times5$.
The size of the door is $1\times1$, which is the same as the agent.
Fetching keys gives 2 rewards to the agent and getting to the goal will give the agent 10 rewards.
No any other rewards.
So it's a hard environment because of narrow doors, long horizon, and sparse reward.
And the difficulty grows up with the increase in the number of keys.

\begin{figure}[ht]
\centering
\includegraphics[width=\linewidth]{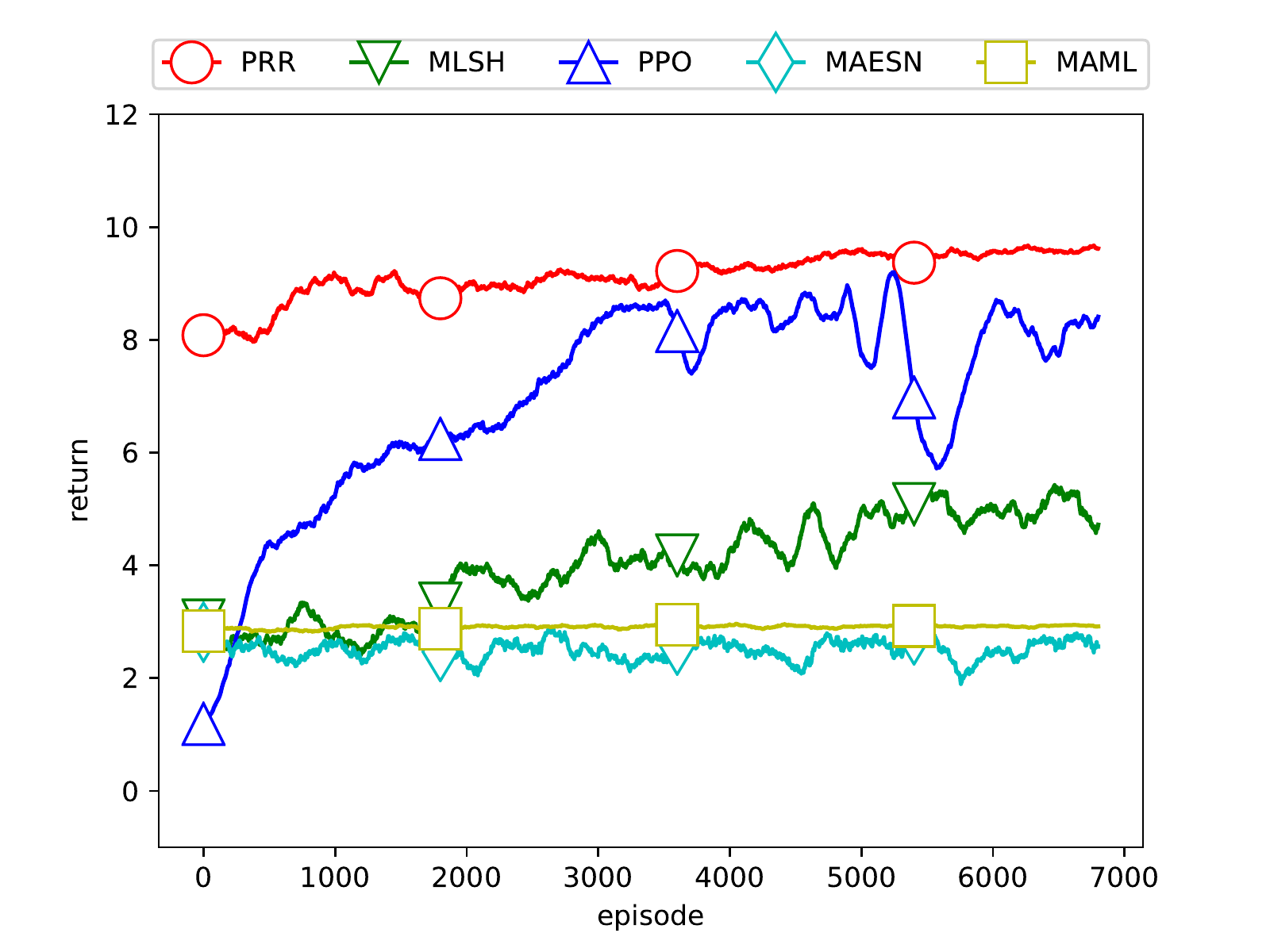}
\caption{Learning curves on key=3 environment.}
\label{fig:keyEnvComp}
\end{figure}

\subsubsection{Overall performance}
In this experiment, we set key=1 and key=2 environments as known tasks.
We want to train PRR on these tasks and then adapt to key=3 environment.

As mentioned before, we first scale the total rewards of all environment to 10.
PPO, as a baseline algorithm, learns on key=3 environment from scratch for 7000 episodes.
MLSH sets four skills and first learns skills on key=1 environment and key=2 environment for 8500 episodes.
And then MLSH transfer to key=3 environment for 7000 training episodes, which the same training episodes as basic PPO baseline.
MAML and MAESN also have 8500 episodes for their meta-training phase on key=1 and key=2 environment.
And then they do meta-testing phase on key=3 environment for 7000 episodes.
PRR have similar experiments setting with other approaches.
It first sets $m_0=[1,1]$ and learns $L_0$ on key=1 environment and key=2 environment for 3000 episodes.
Then PRR sets $m_{11}=[1,0]$ and $m_{12}=[0,1]$, which trains $L_{11}$ on key=1 environment for 1500 training episodes and trains $L_{12}$ on key=2 environment for 4000 episodes.
So PRR totally learns 8500 episodes in its training phase, which is the same as other approaches.
Finally it transfers to key=3 environment for 7000 training episodes and trains $L_{13}$ with $m_{13}=[0,0,1]$.

The learning curves on key=3 environment are shown as Figure \ref{fig:keyEnvComp}.
This environment is quite difficult because of the long horizon and sparse reward.
So the PPO is hard to converge and can not achieve the maximal return.
Our method has a very high return at the beginning and goes up to maximal return.
It indicates that PRR can extract multiple granularities of experience and select proper modules to reuse in a new task, which leads to a good beginning and better exploration.
The MLSH has better a beginning than original PPO.
However, its learning process is very slow and finally reach the return around 5.
MLSH learns fixed granularity of experience and doesn't add new skill when meeting a new environment. Previous skills may not enough to solve the new environment.
MAML and MAESN fail to achieve good results.

\begin{figure*}[!t]
\centering
\includegraphics[width=2in]{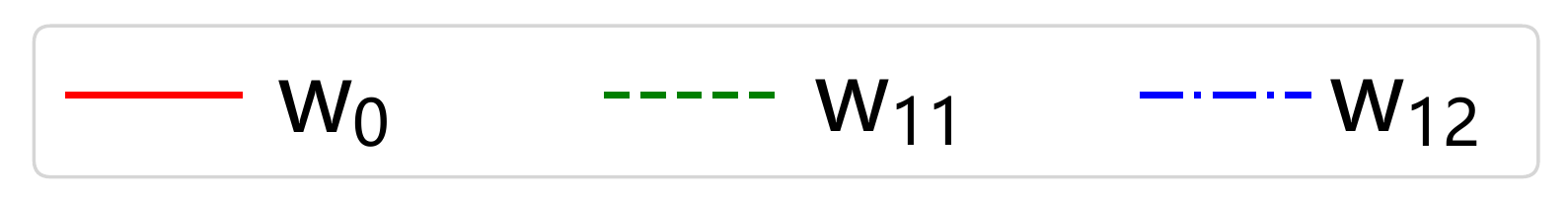}\\
\vspace{0.1in}
\subfigure[Learning $L_{11}$]{
\includegraphics[width=0.23\linewidth]{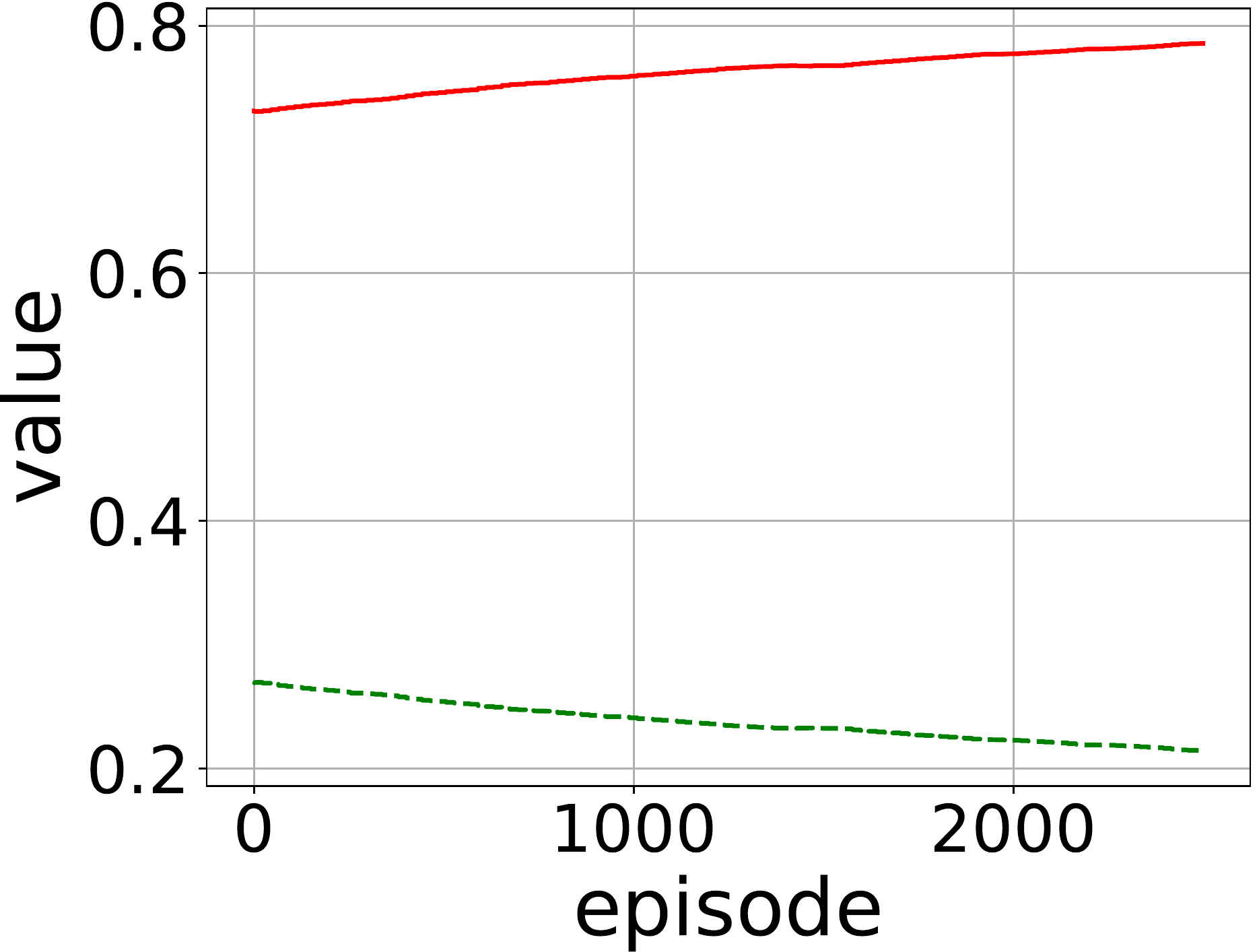}
}
\subfigure[Learning $L_{12}$]{
\includegraphics[width=0.23\linewidth]{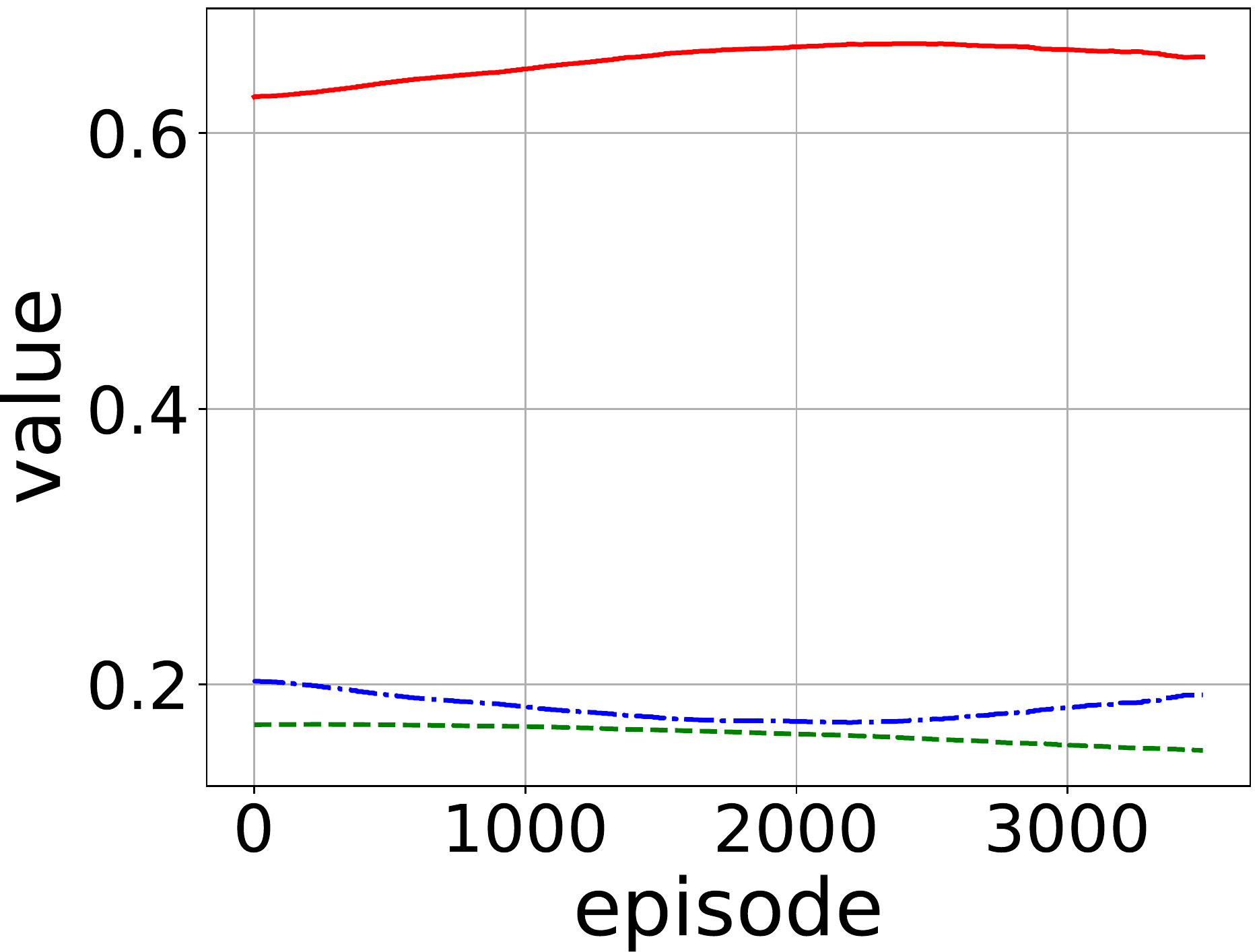}
}
\subfigure[Learning curves on key=3 environment by optimize weights only]{
\includegraphics[width=0.23\linewidth]{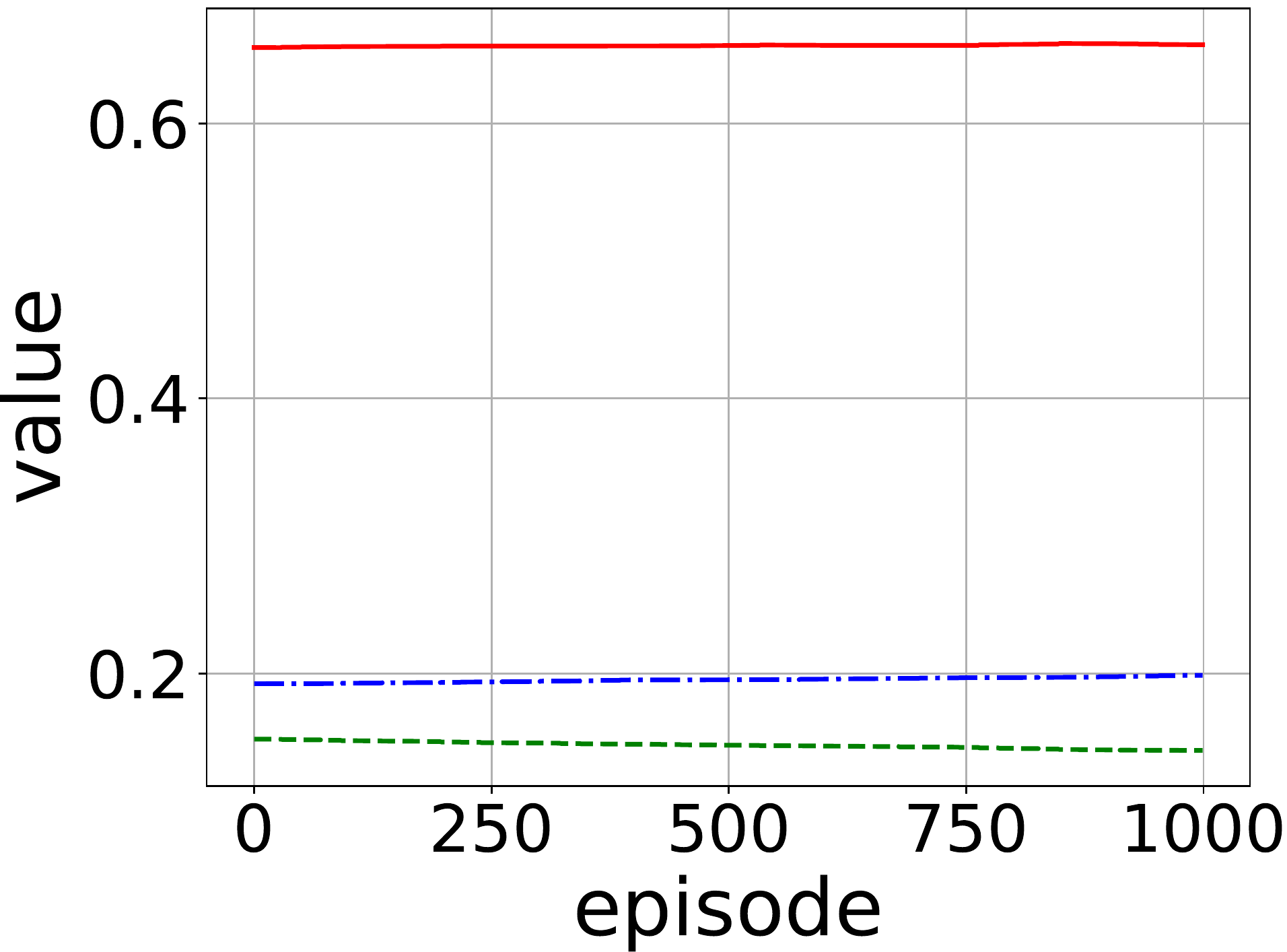}
}
\subfigure[Overall training process]{
\includegraphics[width=0.23\linewidth]{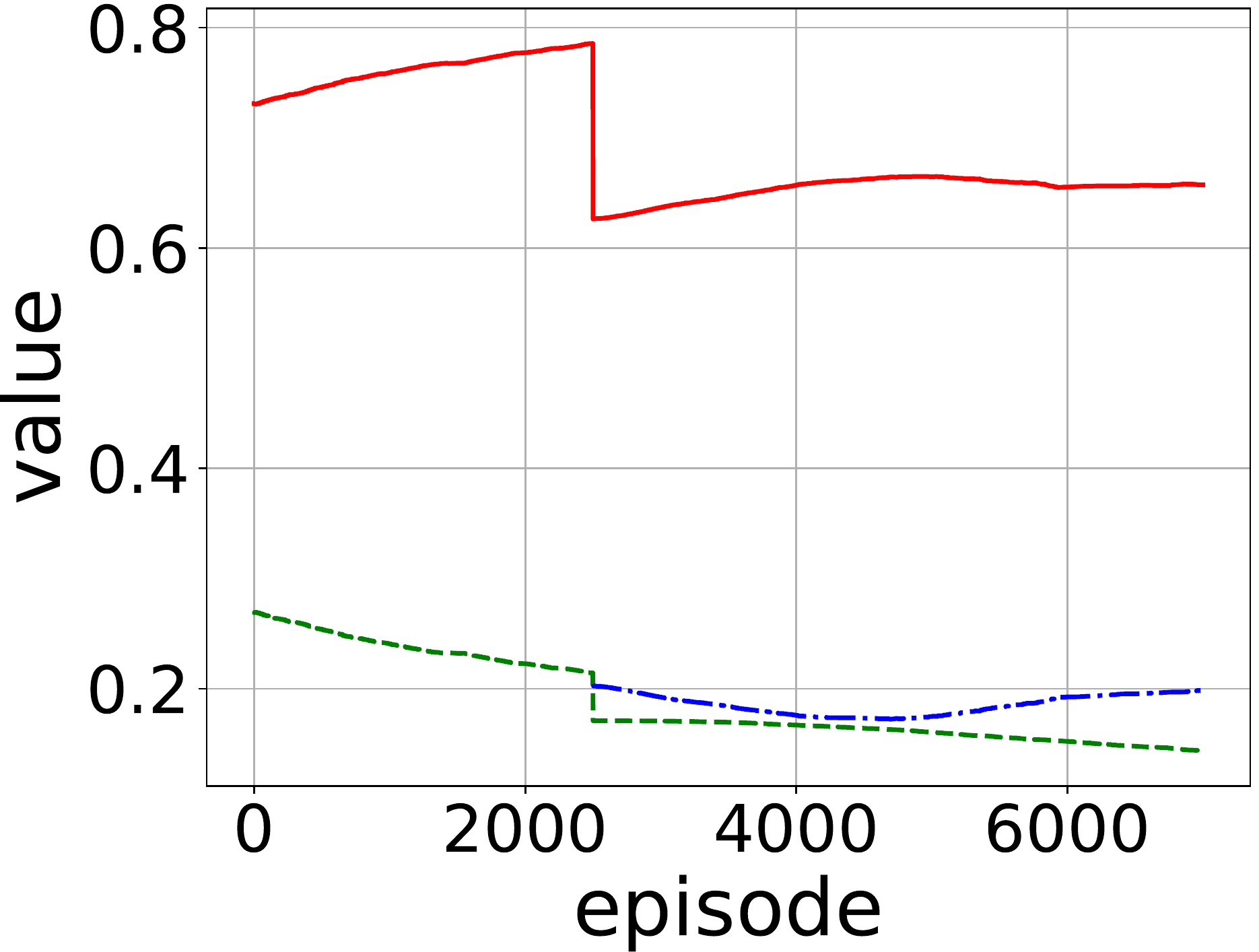}
}
\caption{Weights during training process}
\label{fig:weight_analysis}
\end{figure*}

To make a further analysis of PRR, we look into its learning process.
MLSH also have multi-level architecture and we compare with it.
They are both trained on key=1 task and key=2 task and then reuse experience to learn key=3 task.
As shown in Figure \ref{fig:keyEnvLearnProcess}, PRR learns $L_0$ to extract coarse-grained experience and achieve a good return.
MLSH learns slower than PRR but it still achieves a good result at round return 9.
And then they are trained in key=1 environment.
PRR reuse the experience in $L_0$ to learn $L_{11}$ and the return is 10 all the time.
MLSH learns fixed granularity of experience and doesn't learn new policy, it learns a new high-level policy and fine-tuning its low-level skills.
Then they turn to key=2 environment.
PRR still have a good beginning at return 9 and keep going up by adding a new module $L_{12}$ to fine-grained, low-level experience on key=2 environment.
However MLSH goes up slowly, which is shown in Figure \ref{fig:keyEnvLearnProcess}(c).
The gap is further widened when transferring to key=3 environment.
It shows that fixed granularity of experience can't handle new tasks that dissimilar to old tasks.

\begin{figure}[ht]
\centering
\includegraphics[width=\linewidth]{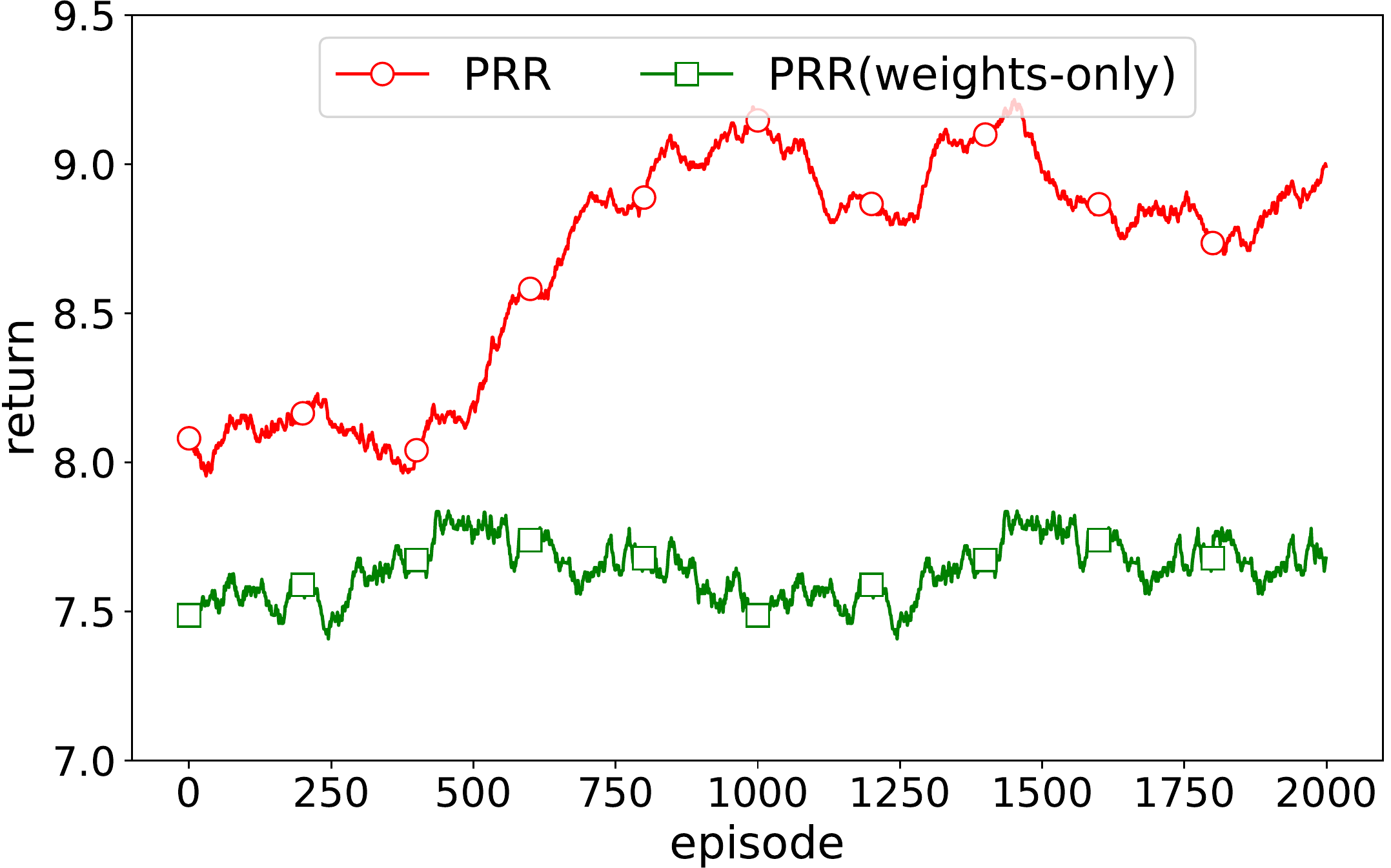}
\caption{Learning curves on key=3 environment by PRR(weights-only) and PRR}
\label{fig:weight_transfer}
\end{figure}

\subsubsection{Analysis of experience selecting}
In order to figure out how does PRR select experience, we look into the weights $\bm{w}$ change during the learning process.
And we also wonder if it's essential to add new networks for key=3 environment.
So we let PRR transfer to the new environment by optimizing weights $\bm{w}$ only and analyze its performance.

In this experiment, we train $L_0$ on key=1 and key=2 environment for 3000 episodes.
Then learn $L_{11}$ and $L_{12}$ on key=1 environment and key=2 environment respectively for 1000 episodes and 3500 episodes, we will look into the change of weights during this process.
And when PRR network transferring to key=3 environment, we don't add new networks and only train weights $\bm{w}$ to mix $L_0$, $L_{11}$ and $L_{12}$.

The change of weights is shown in Figure \ref{fig:weight_analysis}, except for the learning process of $L_0$.
There is only one network when learning $L_0$, the $w_0$ is always 1 during this process.
When learning $L_{11}$, $w_0$ goes up and $w_{11}$ goes down, which as shown in Figure \ref{fig:weight_analysis}(a).
Because key=1 environment is easy so that $L_0$ can handle it well and no more experience should be learned for $L_{11}$.
Then PRR trains $L_{12}$ on key=2 task.
New weight $w_{12}$ are added with initial value 0.
It grows down first and raises again.
It indicates that PRR chooses to use old experience at the beginning because $L_{12}$ is not well trained.
And with the training of $w_{12}$, PPR begins to choose using more $L_{12}$ to solve this task.
Finally we transfer to key=3 environment using $L_0$, $L_{11}$ and $L_{12}$ by optimizing $\bm{w}$.
The weights change very slow during the learning process, which indicates learned experience are not enough to solve this hard new task.
The change of weights during the whole process is shown in Figure \ref{fig:weight_analysis}(d).
The value curve in Figure \ref{fig:weight_analysis}(d) is not smooth, because we normalize $\bm{w}$ during training to ensure that the output of the network is meaningful.

To analyze the performance of this kind of model reuse process, we plot the learning curve of weight-only transfer, which is called PRR (weights-only) and compare to original PRR method, which is shown in Figure \ref{fig:weight_transfer}.
PRR (weights-only) begins at a good point around 7.5.
The return actually goes up by optimizing weights only.
However, it grows very slow and can hardly reach the optimal return at 10.
The result indicates that there are gaps between optimal policy and learned policy residual representations.
This can not be easily filled by adjusting the weights.
So it's essential to learn a new policy residual for a new environment to extract more experience.

\subsubsection{Compare with $L_0$-only learning}
Another issue to be investigated is whether the multiple levels in PPR is really helpful, or it could just the same as the single level architecture (i.e., $L_0$ only).
In this section, we validate the necessity of the proposed residual learning architecture.

We still want to transfer from key=1 and key=2 environments to key=3 environment.
We learn $L_0$ on key=1 and key=2 environments for 3000 episodes.
And then transfer $L_0$ to key=3 environment directly by learning on this environment for 7000 episodes.
This process is like course learning and we denote it as PRR ($L_0$-only) methods.
To compare with this method, origin PRR also learns $L_0$ on key=1 and key=2 environments for the same episodes.
But the full PRR, moreover, learns the residuals on key=1 and key=2 environments, i.e., $L_{11}$ and $L_{12}$ and learn a new policy residual on key=3 environment.

\begin{figure}[ht]
\centering
\includegraphics[width=\linewidth]{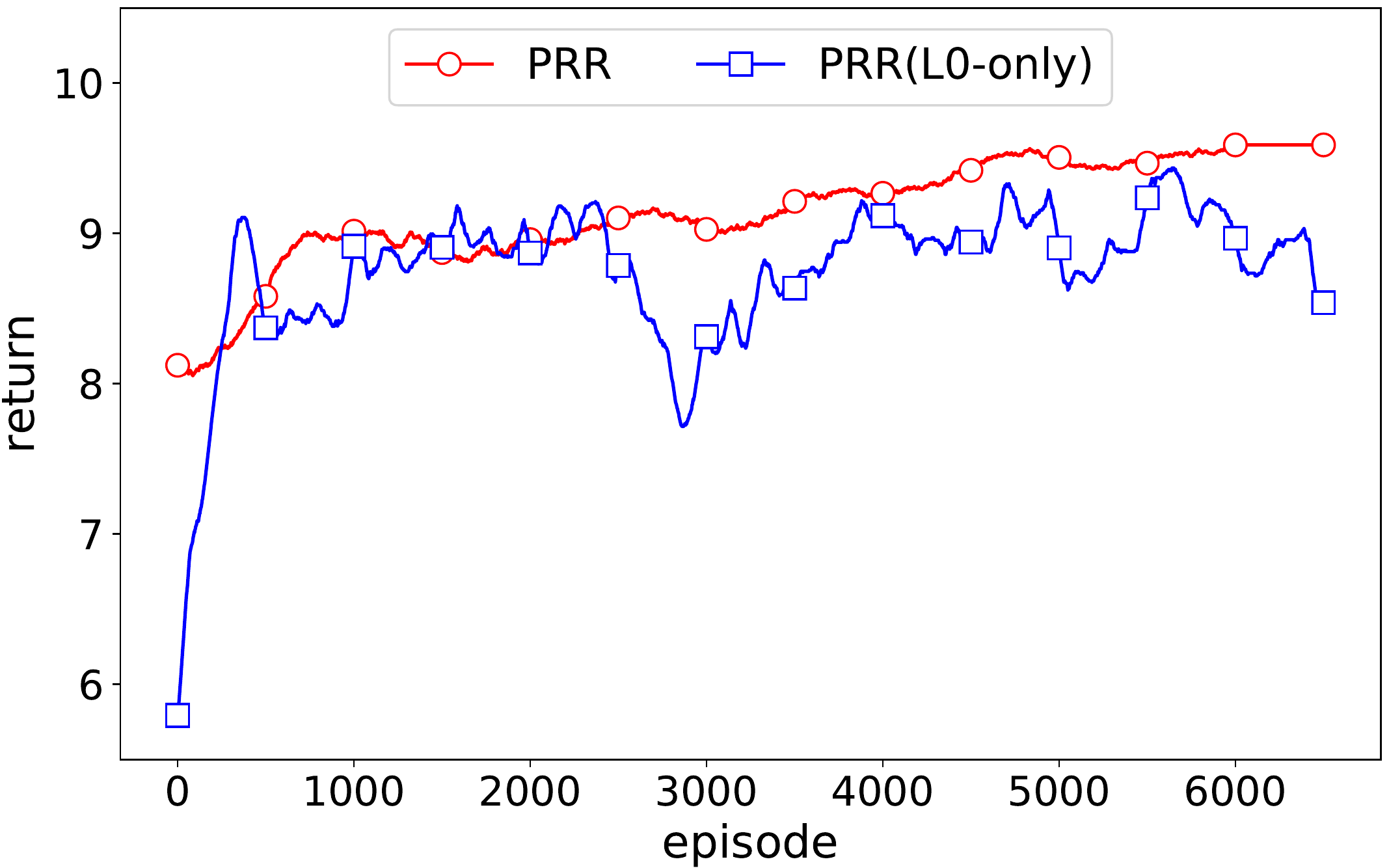}
\caption{Transfer to key=3 environment by PRR($L_0$-only) and PRR}
\label{fig:$L_0$-only}
\end{figure}

The result is shown in Figure \ref{fig:$L_0$-only}.
PRR ($L_0$-only) starts at a lower point than PRR on key=3 environment.
And then it closes to PRR rapidly after some beginning episodes. The rapid grow could due to that it has fewer parameters to learn.
However, PRR ($L_0$-only) does not converge to perform as good as the PRR and shows an unstable learning curve.

The result shows that although a single policy network with previous experience leads to a quick learning process, it still can't capture different granularities of experience in a set of environments.
PRR learns from different levels of experience leads to more comprehensive capturing of environments characteristics.
And this will result in better performance.
Therefore, learning different granularities of experience in different environments is essential.

\subsection{Mujoco}
\begin{figure}[ht]
\centering
\includegraphics[width=0.7\linewidth]{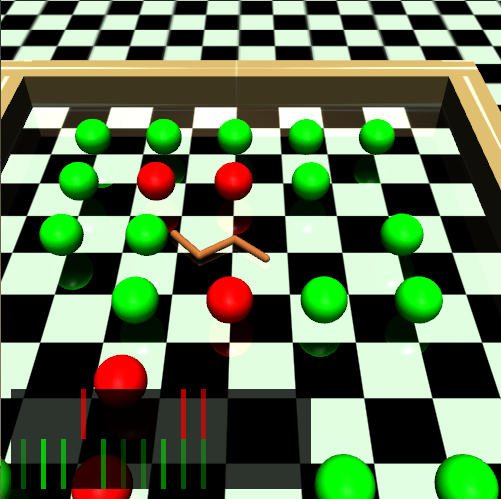}
\caption{SwimmerGather environment}
\label{fig:SwimmerGather}
\end{figure}

We also validate our approach on robot control tasks.
SwimmerGather environment is a hard robot control tasks on Mujoco physics engine \cite{TodorovET12}.
As shown in Figure \ref{fig:SwimmerGather}, the agent is a snake and moves by twisting its body.
The states are observed by sensors and the actions are the angles of its body joint.
The agent gets a reward of 1 for collecting green balls and a reward of -1 for collecting the red ones and no other rewards.
Thus it's a hard robot control environment due to sparse reward and long horizon.

We set three environments.
The first environment has 22 green balls and 0 red balls, which is an easy task.
The second environment has 20 green balls and 3 red balls.
The third environment is the hardest environment that has 18 green balls and 5 red balls.
We first learn $L_0$ by mask $m_{0}=[1,1]$ for 20000 episode.
And learn $L_{11}$ by $m_{11}=[1,0]$ for 2000 episode and learn $L_{12}$ by $m_{12}=[0,1]$ for 2000 episode.
Finally PRR transfers to the third environment and learns $L_{13}$ by $m_{13}=[0,0,1]$.
MLSH has the same number of learning episode as PRR.
PPO, as a baseline algorithm, learns on the third environment from scratch.

\begin{figure}[ht]
\centering
\includegraphics[width=\linewidth]{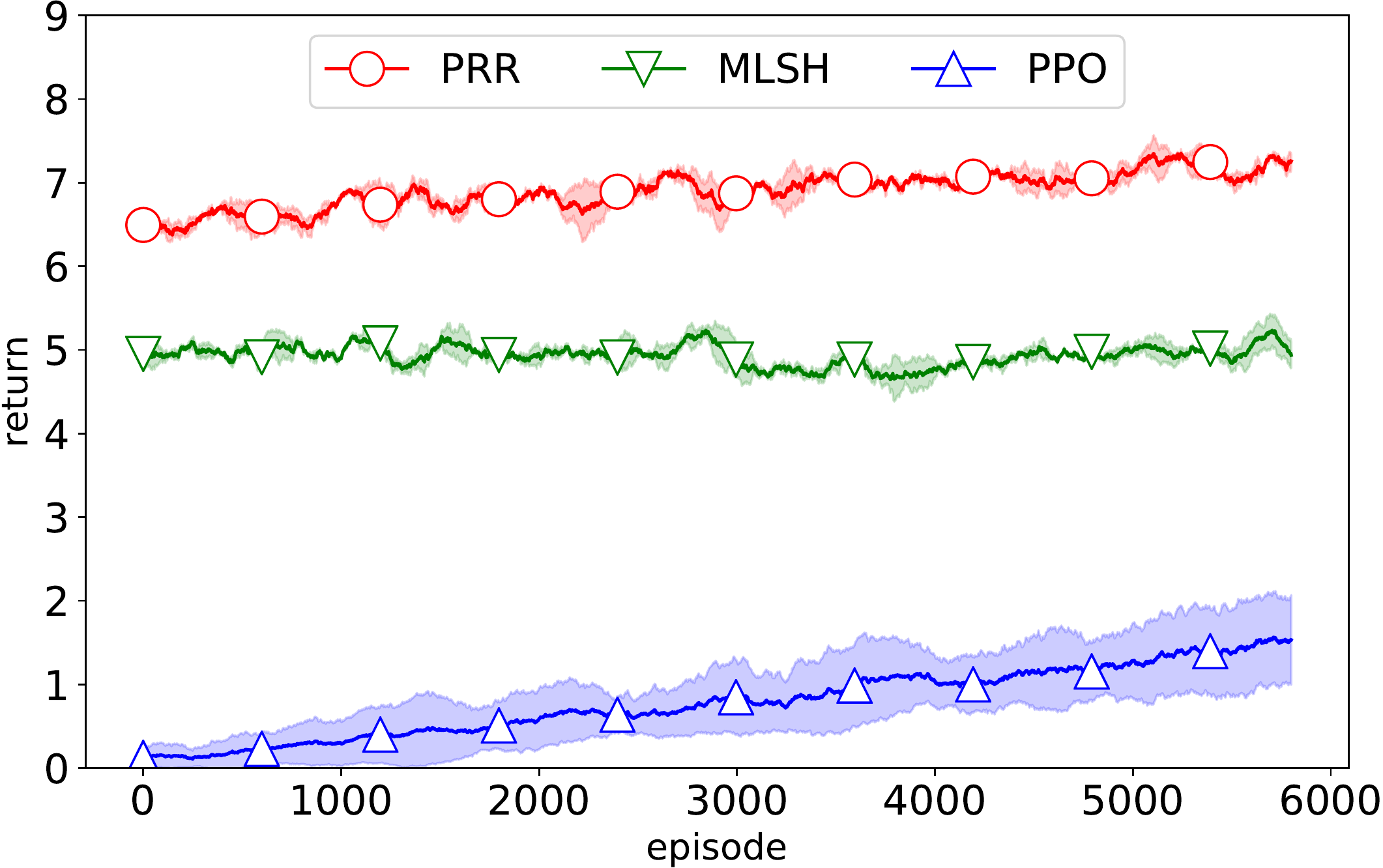}
\caption{Learning curves on SwimmerGather environment}
\label{fig:SwimmerGatherResult}
\end{figure}

\begin{figure*}[ht]
\centering
\subfigure[Learning $L_0$]{
\includegraphics[width=0.45\linewidth]{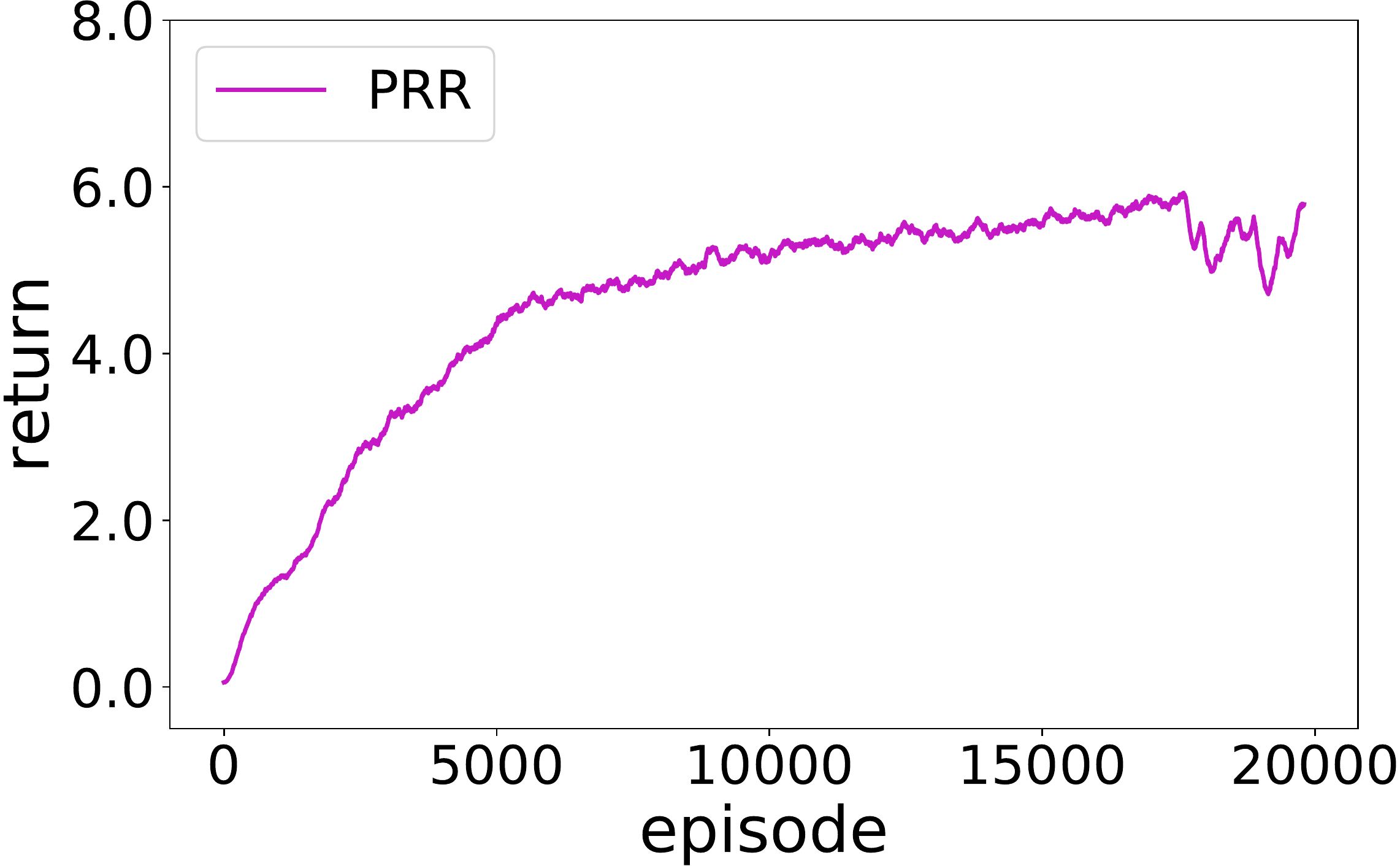}
}
\subfigure[Learning $L_{11}$]{
\includegraphics[width=0.45\linewidth]{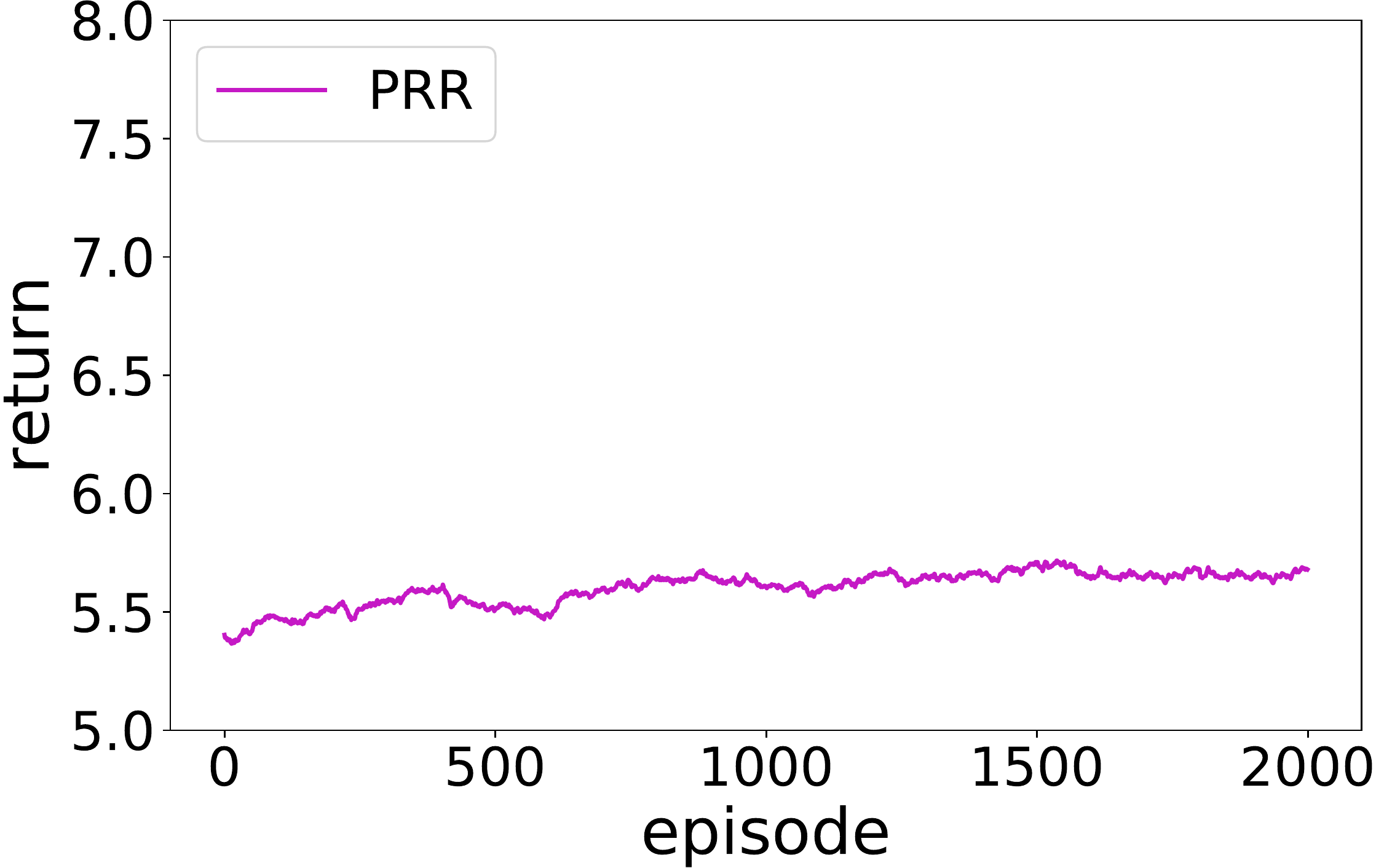}
}
\subfigure[Learning $L_{12}$]{
\includegraphics[width=0.45\linewidth]{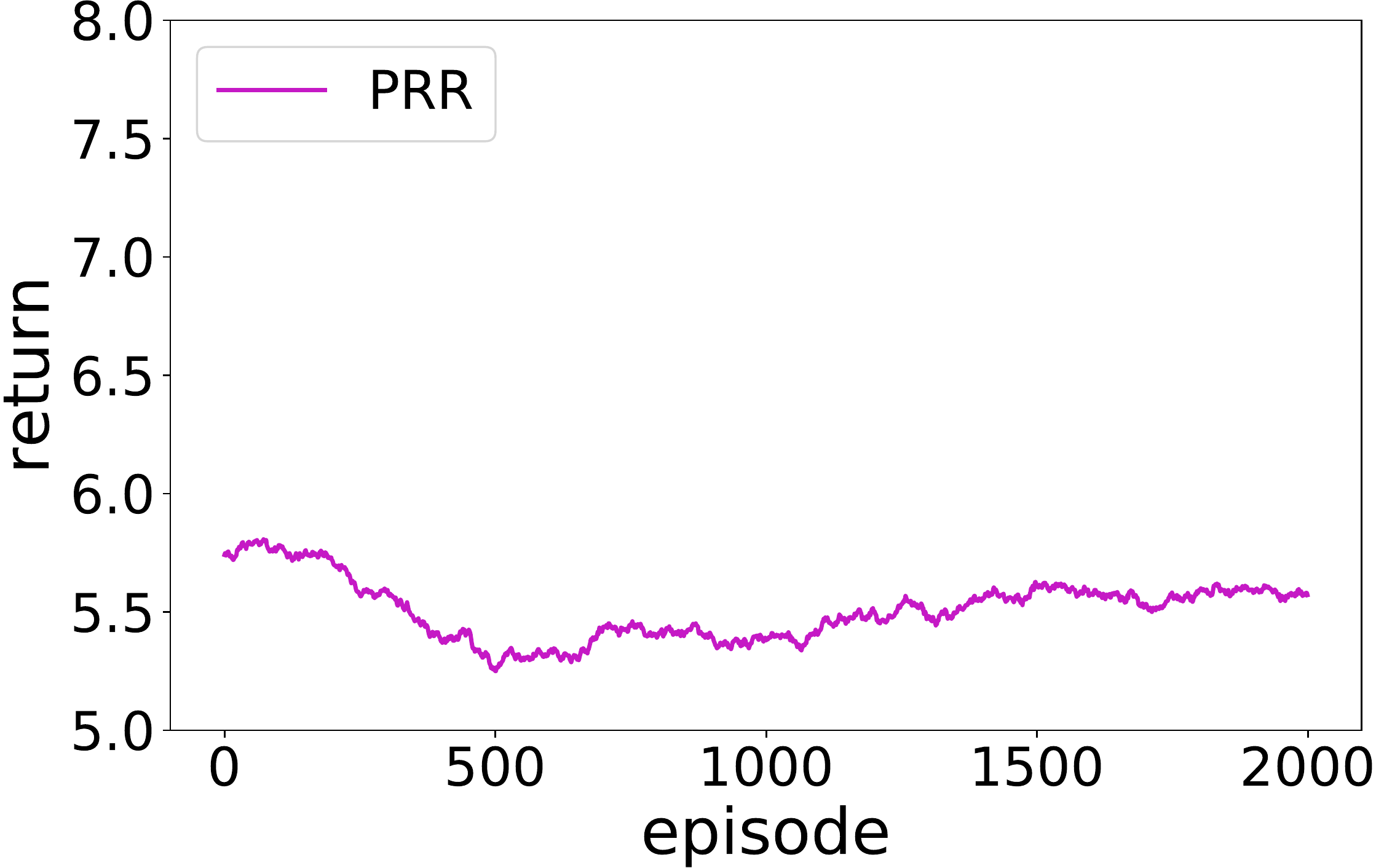}
}
\subfigure[Learning $L_{13}$]{
\includegraphics[width=0.45\linewidth]{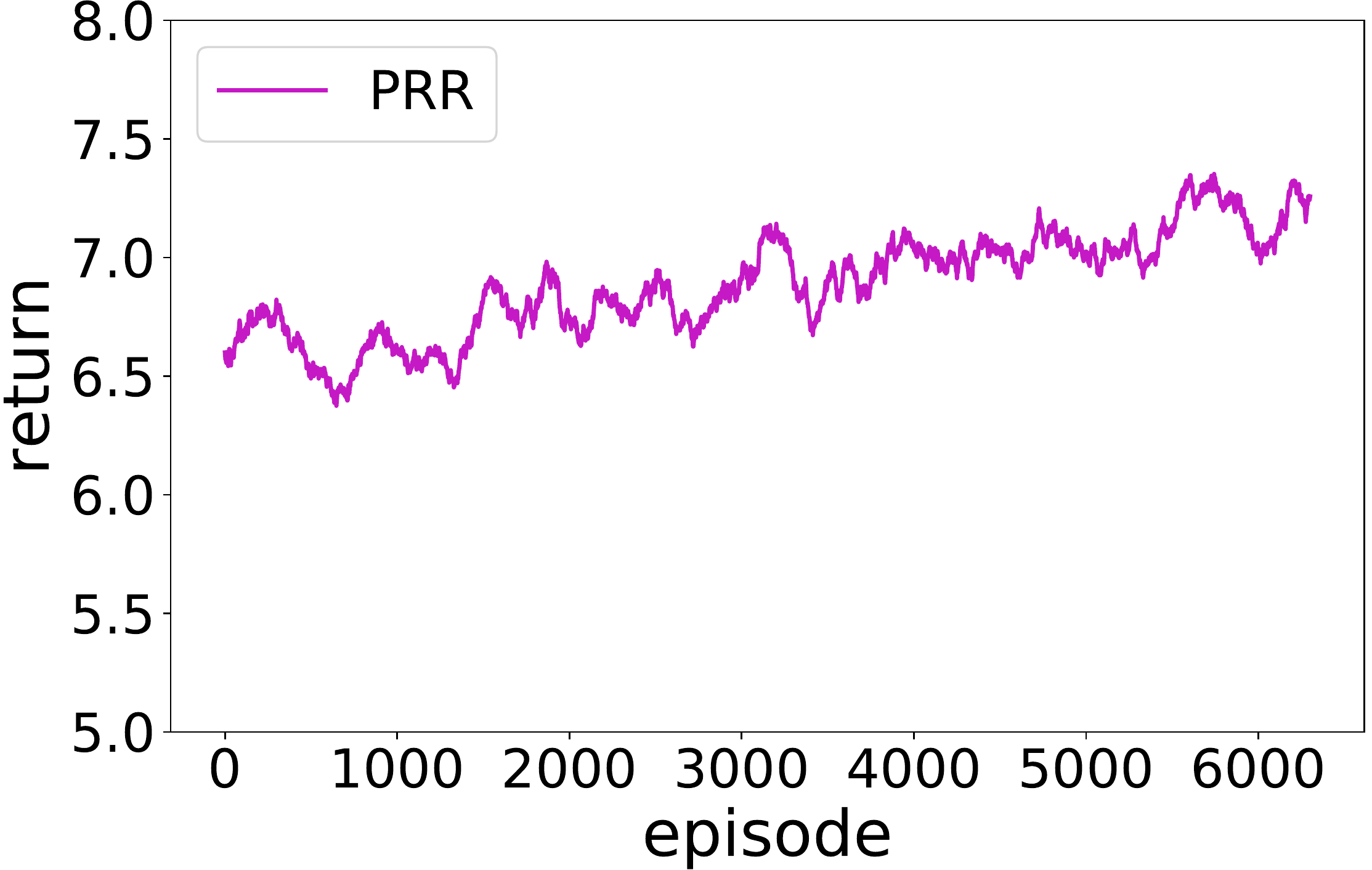}
}
\caption{The training process on SwimmerGather environment.}
\label{fig:SG_learning_process}
\end{figure*}

The learning curves are plotted in Figure \ref{fig:SwimmerGatherResult}.
On this complex environment, learning a policy from scratch is very difficult.
The PPO baseline learns very slow.
In fact, the PPO baseline will reach return at around 4 after 25000 episode.
So it's hard to learn a good policy without transfer from simpler cases.
While both MLSH and PRR use transfer learning, the performance of MLSH is not as good as PRR.
MLSH start at a better point than PPO baseline, but still lower that PRR.
And the return of MLSH grows very slow.
Due to the fixed number of skills, MLSH can hardly draw new experience in new environments, so the return of MLSH grows slowly.
PRR learns different granularities to capture the characteristics of this kind of problems well, so it starts at a very high point in a new environment.
And it learns a new module to fit new knowledge on new environments, which result in visible growth of return in the learning curve.
The learning curves on different levels are shown in Figure \ref{fig:SG_learning_process}.

\begin{figure}[ht!]
\centering
\includegraphics[width=0.8\linewidth]{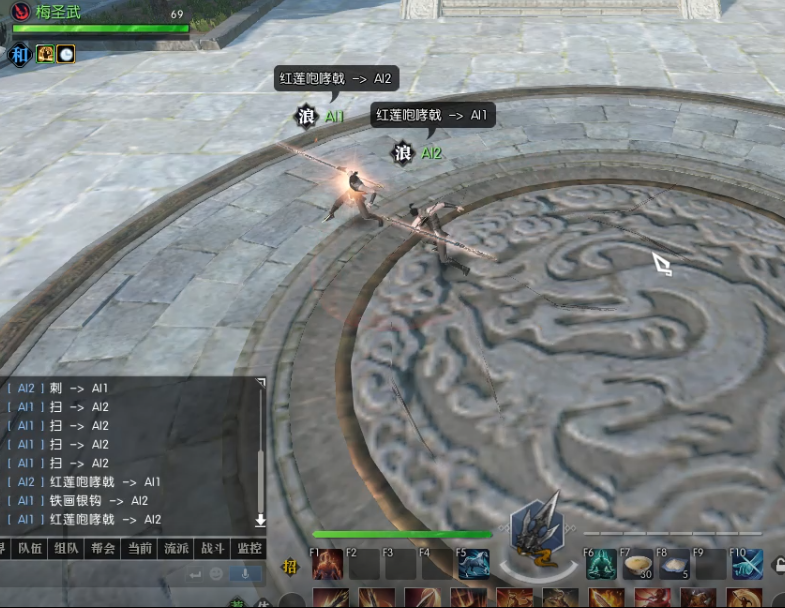}
\caption{Duel mode in NiShuiHan game}
\label{nishuihan}
\end{figure}

\subsection{Experiments on a Fighting Video Game}
We validate our approach on a recent published game NiShuiHan.
As shown in Figure \ref{nishuihan}, it is a one-on-one battle.
There are many roles in this game, training models for each pair of duelists will result in huge time consumption.
So we can apply this approach in this game task, in order to speed up the learning process and reuse the learned model for new roles.

\begin{figure}[ht]
\centering
\includegraphics[width=\linewidth]{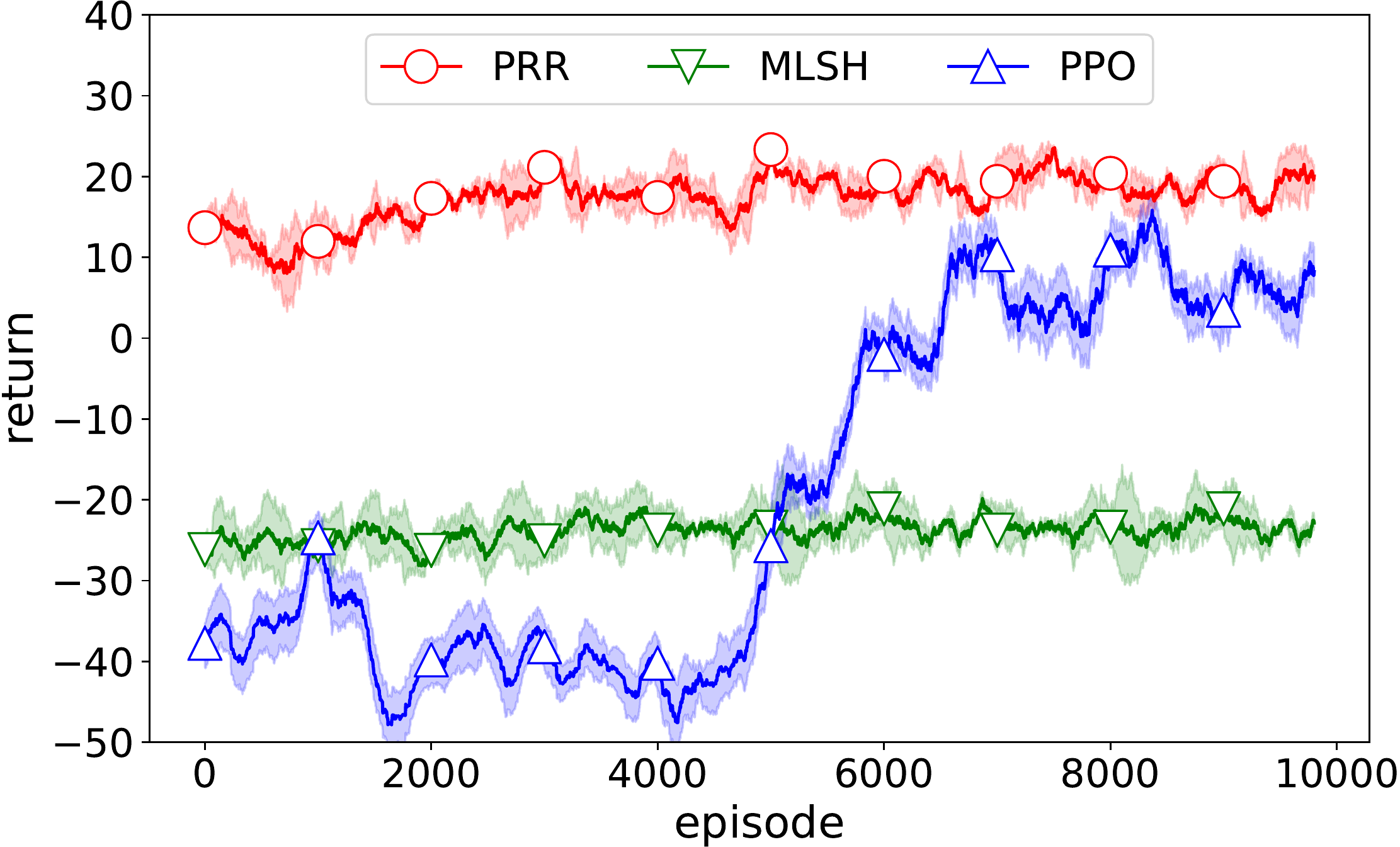}
\caption{Learning curves on video game NiShuiHan}
\label{fig:NSHComp}
\end{figure}

In this experiment, we test on 3 roles, which are denoted as HX, TY, SM.
And we train policy to control HX.
So we have 3 environments, HX v.s. TY, HX v.s. SM and HX v.s. HX.
We first train $L_{0}$, $L_{11}$, $L_{12}$ on HX v.s. TY task and HX v.s. SM task.
And then we transfer it to HX v.s. HX game, which should be the hardest battle in this experiment.
The state is extracted by humans, which is a 101-dimensional vector that contains information about the agent and its opponent.
The action is a 19-dimensional vector that controls agent to move, attack and use skills.
We compare PRR with PPO and MLSH in this experiment.
The result is shown in Figure \ref{fig:NSHComp}.

As shown in Figure \ref{fig:NSHComp}, PPO need a long learning process to learn how to duel with HX.
Both PRR and MLSH have a better beginning, for they both use experience learned from old tasks.
However, MLSH begins at a lower point, and the return almost have no raise during the training process.
PRR begins at a high point, which indicates the different levels of experience can be efficiently reused on a new environment.
And the return still goes up for a better result.
It demonstrates that PRR can efficiently extract experience and reuse them to help training on new tasks.

\section{Conclusion}
This paper addresses the experience presentation issue for experience reuse in reinforcement learning. We propose the policy residual representation (PRR) network that consists of multiple modules organized in levels, and each module is trained on a subset of the tasks to extract the experience in different granularities. We also propose a training process for learning with PRR that trains the modules sequentially, in a top-down level, where a module is trained as the residual policy of the top modules. The experience reuse with a PRR model is then straightforward, by using the PRR model as the initial model with only the final level weights adjustable, leading to a quick adaptation to the new task. For the fine-grained training in the target task, a new module can also be appended to PRR, learned together with the final level weights. Our experiments on 3 sets of tasks show the effectiveness of the PRR model with PPO learning algorithm. Meanwhile, we also notice that the PRR model will become huge when there are a lot of training tasks. Compact PRR models will be studied in the future approaching the goal of learnware \cite{zhou.learnware}.

\bibliographystyle{named}
\bibliography{ijcai19}

\end{document}